\documentclass[10pt,twocolumn,letterpaper]{article}

\usepackage{wacv}
\usepackage{times}
\usepackage{epsfig}
\usepackage{graphicx}
\usepackage{amsmath}
\usepackage{amssymb}

\usepackage[pagebackref=true,breaklinks=true,letterpaper=true,colorlinks,bookmarks=false]{hyperref}

\wacvfinalcopy 

\def\wacvPaperID{436} 

\ifwacvfinal\fi
\begin{document}

\title{Detecting abnormal events in video using Narrowed Normality Clusters}

\author{Radu Tudor Ionescu$^{1,2}$, Sorina Smeureanu$^{1,2}$, Marius Popescu$^{1,2}$, Bogdan Alexe$^{1,2}$\\
$^1$University of Bucharest, 14 Academiei, Bucharest, Romania\\
$^2$SecurifAI, 24 Mircea Vod\u{a}, Bucharest, Romania\vspace*{-0.2cm}
}

\maketitle
\ifwacvfinal\thispagestyle{empty}\fi

\begin{abstract}
We formulate the abnormal event detection problem as an outlier detection task and we propose a two-stage algorithm based on k-means clustering and one-class Support Vector Machines (SVM) to eliminate outliers. In the feature extraction stage, we propose to augment spatio-temporal cubes with deep appearance features extracted from the last convolutional layer of a pre-trained neural network. After extracting motion and appearance features from the training video containing only normal events, we apply k-means clustering to find clusters representing different types of normal motion and appearance features. In the first stage, we consider that clusters with fewer samples (with respect to a given threshold) contain mostly outliers, and we eliminate these clusters altogether. In the second stage, we shrink the borders of the remaining clusters by training a one-class SVM model on each cluster. To detected abnormal events in the test video, we analyze each test sample and consider its maximum normality score provided by the trained one-class SVM models, based on the intuition that a test sample can belong to only one cluster of normality. If the test sample does not fit well in any narrowed normality cluster, then it is labeled as abnormal. We compare our method with several state-of-the-art methods on three benchmark data sets. The empirical results indicate that our abnormal event detection framework can achieve better results in most cases, while processing the test video in real-time at $24$ frames per second on a single CPU.
\end{abstract}

\vspace*{-0.3cm}
\section{Introduction}
\vspace*{-0.1cm}

\begin{figure}[!t]

\begin{center}
\includegraphics[width=1.0\columnwidth]{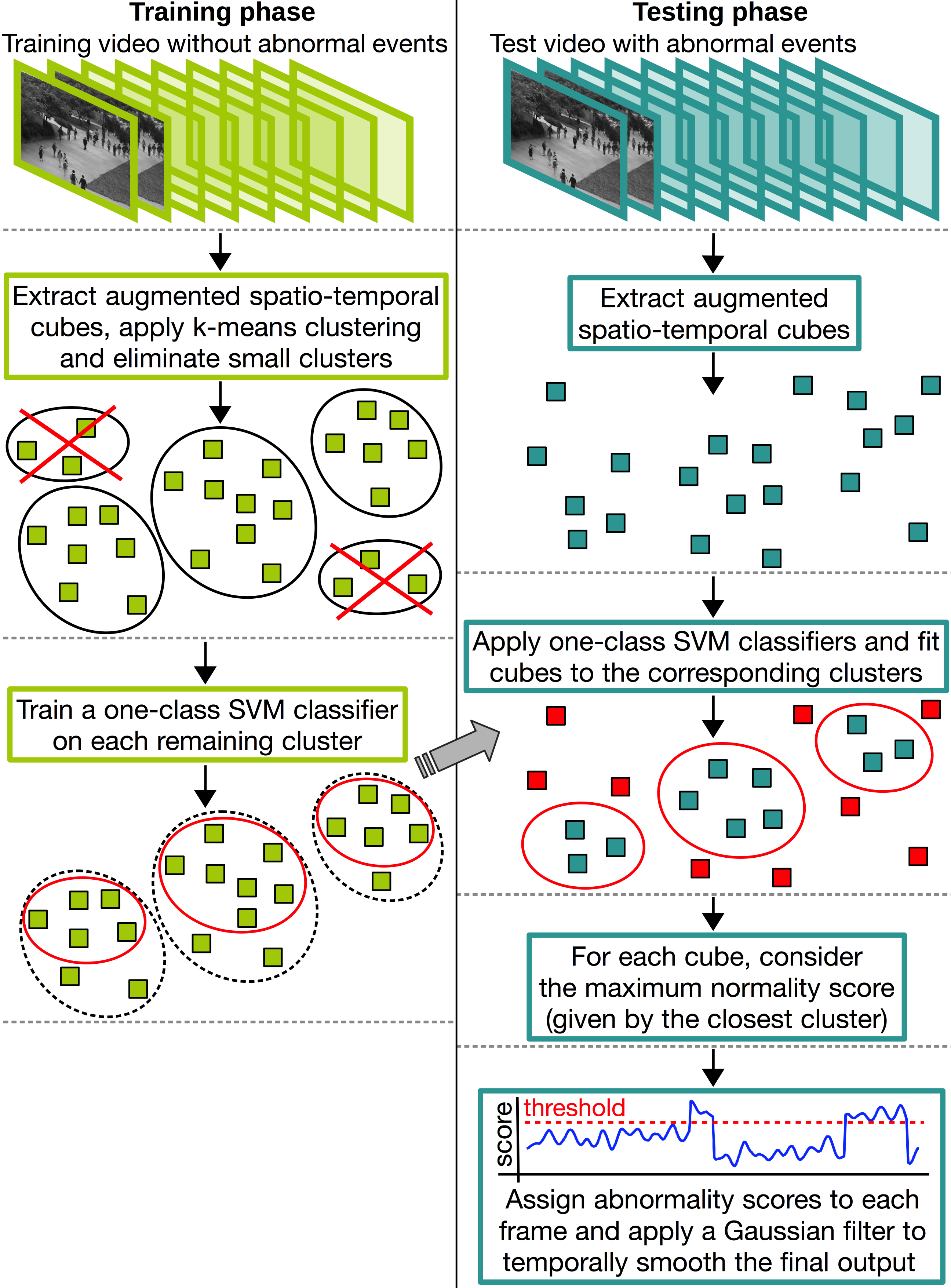}
\end{center}
\vspace*{-0.3cm}
\caption{Our anomaly detection framework based on Narrowed Normality Clusters. In the training phase, we apply a two-stage outlier detection algorithm based on k-means and one-class SVM. In the testing phase, we label a test sample as abnormal if its maximum normality score among the scores provided by the trained one-class SVM models is negative. Best viewed in color.}
\label{fig_pipeline}
\vspace*{-0.45cm}
\end{figure}

Abnormal event detection in video is a challenging task in computer vision, since it is extremely hard, if not impossible, to define abnormal events independent of context. For example, a truck driving by on the street is considered a perfectly normal event, but if the truck drives through a pedestrian area, then it is regarded as an abnormal event. Another example that illustrates the importance of context is represented by two people fighting in a boxing ring (normal event) versus fighting on the street (abnormal event). In addition to the reliance on context, we can generally agree that abnormal behavior should rather be represented by rare (less expected) events~\cite{Itti-CVPR-2005} that do not occur as often as usual (more familiar) events. Due to the scarcity and variability of abnormal events, it is generally hard to obtain a representative set of anomalies at training time. Hence, traditional supervised learning methods are usually ruled out. Therefore, most abnormal event detection approaches~\cite{Antic-ICCV-2011,Cheng-CVPR-2015,Kim-CVPR-2009,Li-PAMI-2014,Lu-ICCV-2013,Mahadevan-CVPR-2010,Mehran-CVPR-2009,Xu-BMVC-2015,Zhao-CVPR-2011} are based on learning a model of normality from training videos containing only familiar events. At test time, events are labeled as abnormal if they deviate from the normality model. We approach abnormal event detection in a similar manner, and propose to build a model of familiarity using the two-stage outlier detection algorithm illustrated in Figure~\ref{fig_pipeline}. We first extract spatio-temporal cubes~\cite{Giorno-ECCV-2016,Ionescu-ICCV-2017,Lu-ICCV-2013}, which we augment with additional information about location, motion direction and object appearance. After extracting augmented spatio-temporal cubes from the training video containing only normal events, we apply k-means clustering to find clusters representing different types of normal motion and appearance. In the first stage, we eliminate the clusters with fewer samples (with respect to a pre-defined threshold), based on the hypothesis that these smaller clusters contain predominantly outliers. Different from other outlier detection approaches based on k-means~\cite{Auslander-SPIE-2011,Chawla-ICDM-2013,Jiang-PRL-2001}, we do not modify the clustering algorithm. Instead, we propose a simple and straightforward approach that aims to coarsely remove some of the outliers, leaving the outliers that are harder to pinpoint for the second stage. In the second stage, we narrow down the borders of the remaining clusters by training a one-class Support Vector Machines (SVM) classifier on each individual cluster. In the end, the learned one-class SVM models represent narrowed clusters of different types of normality. We therefore coin the term \emph{Narrowed Normality Clusters (NNC)} for our two-stage outlier detection algorithm. To detected abnormal events in a test video, we analyze each augmented spatio-temporal cube and consider its maximum normality score among the scores provided by the trained one-class SVM models, based on the natural intuition that a test sample (spatio-temporal cube) should belong to a single narrowed cluster of normal motion and appearance. If the test sample does not fit well in any normality cluster, its corresponding maximum normality score will be negative. Consequently, the respective test sample is labeled as abnormal. 

In summary, the novelty of our paper consists of $(i)$ augmenting spatio-temporal cubes~\cite{Lu-ICCV-2013} with deep appearance features, $(ii)$ assembling together two popular methods for outlier detection (k-means and one-class SVM) into a simple and fast framework and $(iii)$ narrowing down normality clusters by learning a tight border around each cluster.

We conduct experiments on the Avenue \cite{Lu-ICCV-2013}, the Subway \cite{Adam-PAMI-2008} and the UMN \cite{Mehran-CVPR-2009} data sets in order to compare our NNC approach with several state-of-the-art abnormal event detection methods~\cite{Cheng-CVPR-2015,Cong-CVPR-2011,Giorno-ECCV-2016,Hasan-CVPR-2016,Hinami-ICCV-2017,Ionescu-ICCV-2017,Liu-CVPR-2018,Lu-ICCV-2013,Luo-ICCV-2017,Mehran-CVPR-2009,Ravanbakhsh-ICIP-2017,Saligrama-CVPR-2012,Smeureanu-ICIAP-2017,Sun-PR-2017,Zhang-PR-2016}. The empirical results indicate that, on two of the test sets (Avenue and Subway), we obtain better results than all these approaches. We also report the second-best score on the third data set (UMN). It is important to mention that our approach yields impressive results, while running in real-time at $24$ frames per second on a CPU. 

We organize the paper as described next. We present related work on abnormal event detection in Section~\ref{sec_RelatedWork}. We describe our outlier detection framework in Section~\ref{sec_Method}. We present the abnormal event detection experiments in Section~\ref{sec_Experiments}. We draw our final conclusions in Section~\ref{sec_Conclusion}.

\vspace*{-0.2cm}
\section{Related Work}
\label{sec_RelatedWork}
\vspace*{-0.1cm}

Abnormal event detection is commonly formalized as an outlier detection task~\cite{Antic-ICCV-2011,Cheng-CVPR-2015,Cong-CVPR-2011,Dutta-AAAI-2015,Kim-CVPR-2009,Li-PAMI-2014,Lu-ICCV-2013,Mahadevan-CVPR-2010,Mehran-CVPR-2009,Ren-BMVC-2015,Sun-PR-2017,Xu-BMVC-2015,Zhang-PR-2016,Zhao-CVPR-2011}, in which the general approach is to learn a model of normality from training data and label the detected outliers as abnormal events. Some abnormal event detection approaches~\cite{Cheng-CVPR-2015,Cong-CVPR-2011,Dutta-AAAI-2015,Lu-ICCV-2013,Ren-BMVC-2015} are based on learning a dictionary of atoms representing normal events, and on labeling the events not represented in the dictionary as abnormal. At a conceptual level, we can find some resemblance between our approach and dictionary learning. However, going down to the implementation level, there are some important differences. We can interpret the use of k-means to group the training samples into clusters as an unconventional way of building a dictionary of atoms. To the best of our knowledge, there are no dictionary learning approaches that try to remove a part of the atoms as outliers. Unlike dictionary learning approaches, we eliminate the smaller clusters in our framework. Another difference is that we consider that a test sample can belong to a single cluster, or in other words, it can be reconstructed by a single atom. Hence, instead of using the reconstruction error given by a set of basis vectors as the abnormality score, we consider the maximum normality score among the scores given by a set of one-class SVM models, each trained on a different cluster. 

Recent abnormal event detection approaches have employed locality sensitive hashing filters~\cite{Zhang-PR-2016} or deep features~\cite{Hasan-CVPR-2016,Hinami-ICCV-2017,Liu-CVPR-2018,Luo-ICCV-2017,Ravanbakhsh-ICIP-2017,Smeureanu-ICIAP-2017,Xu-BMVC-2015} to achieve better results. Hasan et al.~\cite{Hasan-CVPR-2016} proposed two autoencoders, one that is learned on conventional handcrafted spatio-temporal local features, and another one that is learned end-to-end using a fully convolutional feed-forward architecture. Hinami et al.~\cite{Hinami-ICCV-2017} proposed to train convolutional neural networks (CNN) on multiple visual tasks to exploit semantic information that is useful for detecting and recounting abnormal events, while Smeureanu et al.~\cite{Smeureanu-ICIAP-2017} simply applied convolutional neural networks pre-trained on the ILSVRC benchmark~\cite{Russakovsky2015}. Luo et al.~\cite{Luo-ICCV-2017} proposed a Temporally-coherent Sparse Coding approach, which can be mapped to a stacked Recurrent Neural Network which facilitates parameter optimization and accelerates anomaly prediction. The approach presented in~\cite{Ravanbakhsh-ICIP-2017} is based on training Generative Adversarial Nets (GAN) using normal frames and corresponding optical-flow images in order to learn an internal representation of the scene normality. The test data is compared with both the appearance and the motion representations reconstructed by the GAN and abnormal areas are detected by computing local differences. Liu et al.~\cite{Liu-CVPR-2018} proposed a method for abnormal event detection based on a deep future frame prediction framework. The approach uses the difference between a predicted future frame and the ground-truth frame to detect abnormal events. For a better detection rate, the authors add a temporal constraint based on optical flow along with the spatial constraints.

There have been some approaches that employ unsupervised steps for abnormal event detection~\cite{Dutta-AAAI-2015,Ren-BMVC-2015,Sun-PR-2017,Xu-BMVC-2015}. In~\cite{Dutta-AAAI-2015}, the authors presented a method that constructs a model of familiar events from training data. The model is incrementally updated in an unsupervised manner as new patterns are observed in the test data. In a similar manner, Sun et al.~\cite{Sun-PR-2017} proposed to train a Growing Neural Gas model starting with the training videos and continuing with the test videos, as the test videos are analyzed for anomaly detection. Ren et al.~\cite{Ren-BMVC-2015} used an unsupervised approach, spectral clustering, to construct a dictionary of atoms, each representing a single type of normal behavior. However, the approach of Ren et al.~\cite{Ren-BMVC-2015} requires training videos of normal events to build the dictionary. In order to learn deep feature representations in an unsupervised manner, Xu et al.~\cite{Xu-BMVC-2015} employed Stacked Denoising Auto-Encoders. In the end, they used one-class SVM classifiers to detect the abnormal events. There are some works that do not require any kind of training data in order to detect abnormal events~\cite{Giorno-ECCV-2016,Ionescu-ICCV-2017}. The approach proposed by Del Giorno et al.~\cite{Giorno-ECCV-2016} detects changes in a short sequence of frames from the video by deciding which frames are distinguishable from all the previous frames. Since Del Giorno et al.~\cite{Giorno-ECCV-2016} aimed to obtain an approach independent of temporal ordering, they created shuffles of the test data by permuting the frames before running each instance of the change detection. Ionescu et al.~\cite{Ionescu-ICCV-2017} applied unmasking, a technique based on training a binary classifier to distinguish between two consecutive short video sequences, while gradually removing the most discriminant features. Their hypothesis is that the higher training accuracy rates of the intermediately obtained classifiers represent abnormal events. 

Regarding the feature representation, we use spatio-temporal cubes to represent motion, as other recent approaches~\cite{Giorno-ECCV-2016,Ionescu-ICCV-2017,Lu-ICCV-2013}. Unlike all these approaches, we propose to augment each cube with its location within a spatial pyramid applied over the video frames, with the mean direction given by the 3D motion gradients inside the cube, and with deep appearance features. Our experiments show that the augmentation is useful.
Regarding the outlier detection approach, there are a few works~\cite{Abuolaim-CAIP-2017,Auslander-SPIE-2011} that applied k-means clustering for abnormal event detection. Abuolaim et al.~\cite{Abuolaim-CAIP-2017} used k-means at a coarse level to divide the data points into precisely three clusters: normal, abnormal and ambiguous. On the other hand, we apply k-means with a completely different purpose, namely to obtain many clusters representing different types of normality. Moreover, their approach does not allow to set an abnormality threshold, and thus, it cannot be optimized for better precision or recall. More closely to our approach, Auslander et al.~\cite{Auslander-SPIE-2011} defined three possible assumptions (see Section 4.1 in~\cite{Auslander-SPIE-2011}) for using clustering to detect anomalies. Interestingly, our approach is based on similar assumptions. However, their approach adopts only the first two assumptions defined in~\cite{Auslander-SPIE-2011}, while we satisfy the second assumption by eliminating smaller clusters (in the first stage), and the first and third assumptions by training a one-class SVM on each cluster (in the second stage).

\vspace*{-0.2cm}
\section{Method}
\label{sec_Method}
\vspace*{-0.1cm}

We propose an abnormal event detection framework based on a two-stage algorithm for outlier detection. Our anomaly detection framework is divided into a training phase and a testing phase, as illustrated in Figure~\ref{fig_pipeline}. We next provide an high-level summary of our approach, leaving the additional details about the more important steps for later. From both training and testing videos, we extract spatio-temporal cubes. In the training phase, we cluster the extracted spatio-temporal cubes using k-means and we eliminate the smaller clusters as outliers. On each remaining cluster, we train a one-class SVM model to remove outlier cubes. During inference, each spatio-temporal cube is tested against each one-class SVM model to obtain a set of normality scores. The maximum score is used (with a change of sign) as the abnormality score for the respective test cube. By putting together the cubes from an entire frame, we obtain an anomaly prediction map for each frame. To obtain pixel-level anomaly predictions, the prediction map can be simply resized to match the size of the input video frame. To obtain frame-level predictions, we consider the highest score in the prediction map as the anomaly score of the respective frame. We then apply a Gaussian filter to temporally smooth the final frame-level anomaly scores.


\vspace*{-0.1cm}
\subsection{Feature Extraction}
\label{sec_Features}
\vspace*{-0.1cm}

Unlike other approaches~\cite{Cheng-CVPR-2015,Xu-BMVC-2015}, we apply the same steps in order to extract motion and appearance features from video, irrespective of the data set.

\noindent
{\bf Encoding motion.}
Given the input video, we resize all frames to $120 \times 160$ pixels and uniformly partition each frame to a set of non-overlapping $10 \times 10$ patches. Corresponding patches in $5$ consecutive frames are stacked together to form a spatio-temporal cube. The dimension of each spatio-temporal cube is $10 \times 10 \times 5$. We next derive 3D gradient features from each spatio-temporal cube and normalize the resulted feature vectors using the $L_2$-norm. Until this point, our approach of representing motion is essentially the same as~\cite{Giorno-ECCV-2016,Ionescu-ICCV-2017,Lu-ICCV-2013}. Similar to~\cite{Giorno-ECCV-2016,Ionescu-ICCV-2017,Lu-ICCV-2013}, we eliminate cubes in a region, if the video is static in the respective region. Different from~\cite{Giorno-ECCV-2016,Lu-ICCV-2013}, we do not employ Principal Component Analysis to reduce the feature vector dimension from $500$ to $100$ components, as it has no impact on the performance. Moreover, we diverge from the standard spatio-temporal cube representation by augmenting the cubes with additional information about location, motion direction and object appearance, as described next.

\noindent
{\bf Encoding location.}
We divide each frame into a spatial pyramid~\cite{Lazebnik-BBF-2006} with two levels, the first level containing $2 \times 2$ bins and the second one containing $4 \times 4$ bins. We encode the location of each spatio-temporal cube as a one-hot vector for each level of the pyramid. This gives $20$ additional features ($2 \times 2 + 4 \times 4$) for each cube. The purpose of recording spatial information into the cube representation is to accurately detect situations in which abnormal events can appear in only some region of the video. For instance in a traffic surveillance video, people crossing the street on a crosswalk is a normal event, but if they cross it outside the designated area this should be labeled as abnormal.
 
\noindent
{\bf Encoding mean direction.}
To extract the mean motion direction from each spatio-temporal cube, we first consider the individual patches of the cube. In each patch, we compute the center of mass of the 3D gradients. We then encode the displacement of the center of mass in consecutive patches as vectors representing motion direction. For a better estimation of the mean motion direction, we also compute motion direction vectors after dividing each patch into $2 \times 2$ bins. Finally, the motion direction vectors are quantized into an orientation-based histogram with $8$ bins. The histogram bins are evenly spread over $0$ to $360$ degrees. Our histogram is produced in a similar way to the histogram corresponding to a cell in the Histograms of Oriented Gradients (HOG) descriptor~\cite{Dalal-HOG-2005}. Along with the histogram, we add another feature given by the sum of all vector magnitudes. In total, there are $9$ additional features for augmenting the cube. The purpose of recording the mean direction into the cube representation is to enable the accurate detection of abnormal events triggered by objects moving in a certain direction. For example in a traffic surveillance video, a car driving the wrong way should be labeled as abnormal.

\noindent
{\bf Encoding object appearance.}
In many computer vision tasks, for instance image difficulty prediction~\cite{img-difficulty-CVPR-2016}, 
higher level features, such as the ones learned with convolutional neural networks (CNN)~\cite{Hinton-NIPS-2012} are the most effective. To build our appearance features, we consider a shallow pre-trained CNN architecture, namely VGG-f~\cite{Chatfield-BMVC-14}, which is able to process the video frames in real-time on a CPU. Considering that we want our detection framework to work in real-time on a standard desktop computer, not equipped with expensive GPU, the VGG-f architecture is an excellent choice.
We hereby note that better anomaly detection performance can probably be achieved by employing deeper CNN architectures, such as VGG-verydeep~\cite{Simonyan-ICLR-14} or ResNet~\cite{He-CVPR-2016}.
We use a VGG-f model pre-trained on the ILSVRC benchmark~\cite{Russakovsky2015} to extract deep features as follows. Given the input video, we resize the frames to $224 \times 224$ pixels. We then subtract the mean imagine from each frame and provide it as input to the VGG-f model. We remove the fully-connected layers (identified as \emph{fc6}, \emph{fc7} and \emph{softmax}) and consider the activation maps of the last convolutional layer (\emph{conv5}) as appearance features. While the fully-connected layers are adapted for object recognition, the last convolutional layer contains valuable appearance and pose information which is more useful for our anomaly detection task. Ideally, we would like to have at least slightly different representations for a person walking versus a person running, hence \emph{conv5} is more suitable than \emph{fc6} or \emph{fc7}. From the \emph{conv5} layer, we obtain $256$ activation maps, each of $13 \times 13$ units. We then resize each activation map to match the number of spatio-temporal cubes in a frame, i.e. from $13 \times 13$ to $12 \times 16$, using bicubic interpolation. We then concatenate each set of $256$ filter activations to the corresponding spatio-temporal cube. 

\vspace*{-0.1cm}
\subsection{Two-Stage Outlier Detection}
\vspace*{-0.1cm}

\begin{figure}[!t]
\begin{center}
\includegraphics[width=0.85\columnwidth]{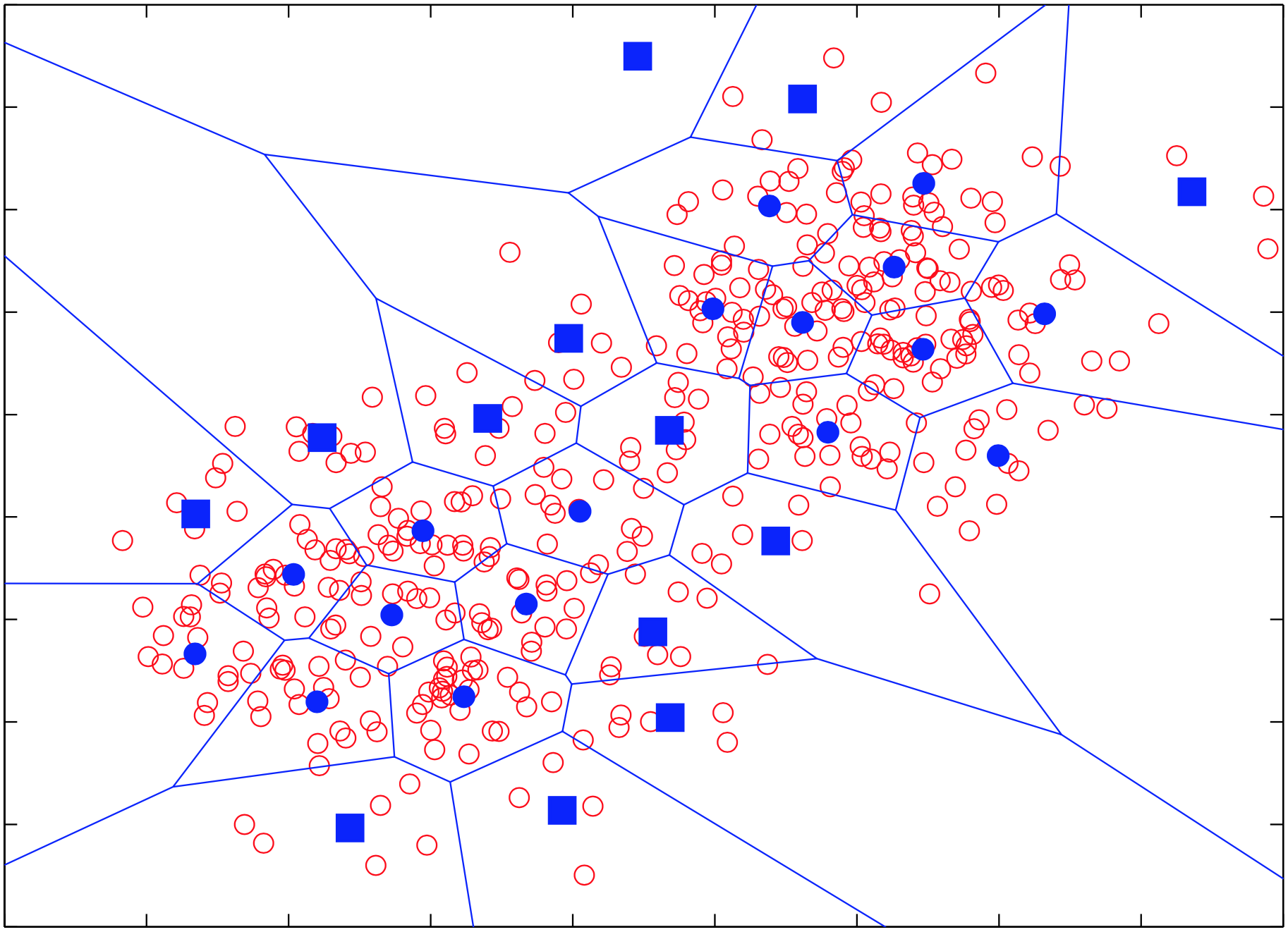}
\end{center}
\vspace*{-0.2cm}
\caption{A set of $400$ data points sampled from two normal distributions of different means. The points are clustered into $30$ clusters using k-means. The centroids of clusters with less than $10$ samples are represented with a large blue square. Best viewed in color.}
\label{fig_kmeans}
\vspace*{-0.4cm}
\end{figure}

\begin{figure}[!t]
\begin{center}
\includegraphics[width=0.68\columnwidth]{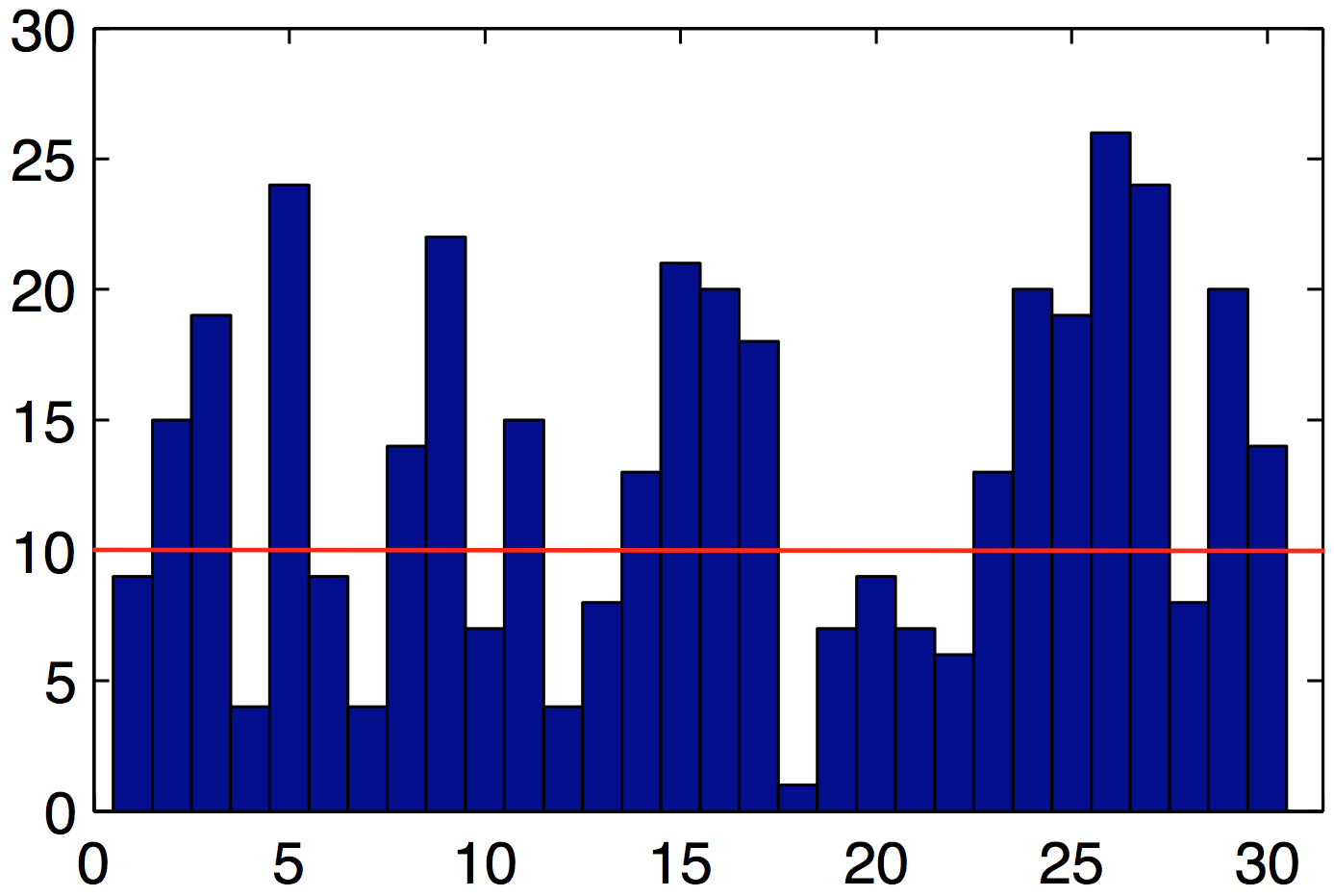}
\end{center}
\vspace*{-0.2cm}
\caption{A histogram representing the number of data points in each cluster. The histogram corresponds to the k-means clustering applied over the $400$ data points illustrated in Figure~\ref{fig_kmeans}. A threshold of $10$ is used to detect clusters of outliers. Best viewed in color.}
\label{fig_hist}
\vspace*{-0.4cm}
\end{figure}

\noindent
{\bf First stage detection based on k-means.}
We cluster the augmented spatio-temporal cubes extracted from the training video to find clusters representing different types of normality. Next, we eliminate the clusters with fewer samples, based on the assumption that these smaller clusters contain mostly outlier samples. We note that the same assumption also sits at the basis of the method proposed in~\cite{Auslander-SPIE-2011}. Nonetheless, we motivate the assumption through the following toy example. We generate $400$ data points sampled from two normal distributions of different means. We group the points into $k=30$ clusters using k-means and we illustrate the result in Figure~\ref{fig_kmeans}. We then count the number of points in each cluster and obtain the histogram depicted in Figure~\ref{fig_hist}. In this example, we consider that the clusters with less than $10$ data points contain mostly outliers. The centroids of these smaller clusters are marked with a large blue square in Figure~\ref{fig_kmeans}. We can clearly see that the marked clusters are farthest from both normal distribution means, indicating that the containing points are indeed outliers. Nevertheless, our aim is to test out the assumption on real data, in the context of abnormal event detection in video. Although the training does not contain abnormal events, we believe that k-means helps to remove noisy or weak patterns that can occur in the normal video. 

\noindent
{\bf Second stage detection based on one-class SVM.}
After removing the smaller k-means clusters, we are left with a set of clusters $\mathcal{C} = \{ c_1, c_2, ...,c_r \,|\, r \leq k\}$ that accurately model the stronger patterns of normality. However, k-means does not provide a tight boundary around the remaining clusters, and, in some cases, it leaves a lot of room to accommodate outliers. For example, the borders of some remaining clusters represented in Figure~\ref{fig_kmeans} span to infinity. To alleviate this problem, we propose to narrow down the borders of the remaining clusters by training a one-class SVM~\cite{Scholkopf-2001} classifier on each cluster. We note that the border learned by SVM is tighter (or narrower) than the border of the original cluster (which includes all cluster's samples), since the one-class SVM model is forced to single out a small percentage of samples within the cluster as outliers. In this regard, we can state that one-class SVM narrows (or tightens) the border around the cluster's centroid. Hence, the learned one-class SVM models can be interpreted as a set of narrowed clusters representing different types of normal motion and appearance. To train our set of classifiers, we consider each spatio-temporal cube as an independent and individual sample, disregarding the temporal relations among cubes. Let $\mathcal{X} = \{ x_1, x_2, ...,x_n \,|\, x_i \in \mathbb{R}^m\}$ denote the set of training cubes in a given cluster $c_j$. The one-class SVM model corresponding to a cluster $c_j$ will learn to separate a small region capturing the normal cubes inside cluster $c_j$ from the rest of the feature space, by maximizing the distance from the separating hyperplane to the origin. This associates to each cluster a binary classification function $g$ that singles out a region in the input space where the probability density of a particular type of normality lives:
\begin{equation}\label{eq_1}
g(x) = sign \left( \sum_{i=1}^{n} \alpha_i k(x, x_i) - \rho \right),
\end{equation}
where $x$ is a test cube that must be classified either as normal or abnormal, $x_i \in \mathcal{X}$ is a training cube, $\alpha_i$ are the weights assigned to the support vectors $x_i$, $\rho$ is the distance from the hyperplane to the origin, and $k$ is a kernel function, in our case, the linear kernel. If we just need a score reflecting the normality level of a spatio-temporal cube, we can simply remove the sign transfer function from Equation~\eqref{eq_1} and obtain a scoring function. It is important to note that for each cluster $c_j \in \mathcal{C}$, we have a different scoring function $g_{c_j}$. Then, for a given test cube, we will have a set of $r$ normality scores. However, since the narrowed clusters are independent (they reside in different areas of the feature space), we can naturally assume that a spatio-temporal cube belongs to a single cluster. Therefore, we consider the maximum normality score, the one that corresponds to the narrowed cluster that better fits (is closer to) the test cube. If the test spatio-temporal cube does not fit well in any normality cluster, its corresponding maximum normality score will be negative (the cube is outside the nearest cluster). Consequently, the respective test sample is labeled as abnormal.


\vspace*{-0.2cm}
\section{Experiments}
\label{sec_Experiments}

\vspace*{-0.1cm}
\subsection{Data Sets}
\vspace*{-0.1cm}

We consider three data sets for the abnormal event detection experiments.

\noindent
{\bf Avenue.}
The Avenue data set~\cite{Lu-ICCV-2013} is composed of $16$ training videos and $21$ test videos. In total, the Avenue data set contains $15328$ frames for training and $15324$ frames for testing. The resolution of each frame is $360 \times 640$ pixels. The locations of anomalies are annotated in ground-truth pixel-level masks for each frame in the test videos. Hinami et al.~\cite{Hinami-ICCV-2017} argued that the Avenue test set contains five videos ($1$, $2$, $8$, $9$ and $10$) with static abnormal objects that are not properly annotated. Hence, they evaluated their approach on a subset (Avenue17) that excludes these five videos. When we compare our results with those reported in~\cite{Hinami-ICCV-2017}, we also remove the same five videos for a fair comparison.

\noindent
{\bf Subway.}
The Subway surveillance data set~\cite{Adam-PAMI-2008} is one of the largest data sets for anomaly detection in video. The Subway data set is formed of two videos, one of $96$ minutes (Entrance gate) and another one of $43$ minutes (Exit gate). The Entrance gate video contains $144251$ frames, while the Exit gate video contains $64903$ frames. The resolution of each video frame is $384 \times 512$ pixels. Abnormal events are labeled at the frame level. 

\noindent
{\bf UMN.}
The UMN Unusual Crowd Activity data set~\cite{Mehran-CVPR-2009} is composed of three different crowded scenes of various lengths. The first scene contains $1453$ frames, the second scene contains $4144$ frames, and the third scene contains $2144$ frames. The resolution of each frame is $240 \times 320$ pixels. In the normal scenario, people walk around in the scene, and the abnormal behavior is defined as people running in all directions.

\vspace*{-0.1cm}
\subsection{Evaluation}
\vspace*{-0.1cm}

As evaluation metrics, we opt for ROC curves and the corresponding \emph{area under the curve} (AUC) computed with respect to ground-truth frame-level annotations, and, when available (for the Avenue data set), pixel-level annotations. We use the same frame-level and pixel-level AUC definitions as in previous works~\cite{Cong-CVPR-2011,Giorno-ECCV-2016,Ionescu-ICCV-2017,Lu-ICCV-2013,Mahadevan-CVPR-2010,Sun-PR-2017,Xu-BMVC-2015}. At the frame-level, a frame is considered a correct detection if it contains at least one abnormal pixel. At the pixel-level, the corresponding frame is considered as being correctly detected if more than $40\%$ of truly anomalous pixels are detected. Before the evaluation, we smooth the pixel-level detection maps with the same filter used by~\cite{Giorno-ECCV-2016,Ionescu-ICCV-2017,Lu-ICCV-2013} in order to obtain the final pixel-level detections. We exclude the recent approach (based on k-means) of Abuolaim et al.~\cite{Abuolaim-CAIP-2017} from our evaluation, because their approach is constrained to provide a single point on the ROC curve and the frame-level or pixel-level AUC metrics cannot be determined in their case. Many works~\cite{Cong-CVPR-2011,Dutta-AAAI-2015,Lu-ICCV-2013,Mahadevan-CVPR-2010,Xu-BMVC-2015,Zhang-PR-2016} include the Equal Error Rate (EER) as evaluation metric, but some recent works~\cite{Giorno-ECCV-2016,Ionescu-ICCV-2017} argue that metrics such as the EER can be misleading in a realistic anomaly detection setting, in which abnormal events are expected to be very rare. As we agree with perspective of~\cite{Giorno-ECCV-2016,Ionescu-ICCV-2017}, we refrain from employing the EER in our evaluation.

\vspace*{-0.1cm}
\subsection{Parameter and Implementation Details}
\vspace*{-0.1cm}

We extract spatio-temporal cubes from the training and the test video sequences using the code available online at {https://alliedel.github.io/anomalydetection/}. We use our own implementation to augment the cubes with location and mean direction. For the appearance features, we consider the pre-trained VGG-f~\cite{Chatfield-BMVC-14} model provided in MatConvNet~\cite{matconvnet}, and extract the features from the \emph{conv5} layer. For a faster processing time, we extract features for one in every two frames in the test video, without observing any drop in performance. To cluster the augmented cubes, we employ the k-means implementation from VLFeat~\cite{vedaldi-vlfeat-2008} based on the original Lloyd algorithm~\cite{Du-SIAM-1999}. We use k-means++~\cite{Arthur-SODA-2007} initialization. We repeat the clustering $10$ times and choose the partitioning with the minimum energy. We choose the number of clusters $k$ such that we have on average $1000$ cubes per cluster, hence $k$ is always proportional to the number of training cubes. For instance, there are almost $280$ thousand cubes extracted from the Avenue training videos, hence we set $k=280$ for the Avenue data set. We then eliminate the clusters with less than $500$ cubes, irrespective of the data set. 
To remove the outliers from each cluster, we employ the one-class SVM implementation from LibSVM~\cite{LibSVM-2011}. In all the experiments, we set the regularization parameter of one-class SVM to $0.01$, which means that the model will have to single out $99\%$ of the training cubes as normal (the other $1\%$ are outliers). At test time, we are able to process the test videos at nearly $24$ FPS on a computer with Intel Core i7 2.3 GHz processor and 8 GB of RAM.

\vspace*{-0.1cm}
\subsection{Results on the Avenue Data Set}
\vspace*{-0.1cm}

\begin{table}[t]
\small{
\begin{center}
\begin{tabular}{|l|c|c|}
\hline
Method 																& \multicolumn{2}{|c|}{AUC} \\
\cline{2-3}
			 																& Frame  		& Pixel\\
\hline
\hline
Lu et al.~\cite{Lu-ICCV-2013}								& $80.9$			& $92.9$ \\
Hasan et al.~\cite{Hasan-CVPR-2016}					& $70.2$			& -			\\
Del Giorno et al.~\cite{Giorno-ECCV-2016}			& $78.3$			& $91.0$ \\
Smeureanu et al.~\cite{Smeureanu-ICIAP-2017}	& $84.6$			& $93.5$ \\
Ionescu et al.~\cite{Ionescu-ICCV-2017}				& $80.6$			& $93.0$ \\
Luo et al.~\cite{Luo-ICCV-2017}							& $81.7$			& - 			\\
Liu et al.~\cite{Liu-CVPR-2018}								& $85.1$			& - 			\\
\hline
cubes + one-class SVM  										& $81.3$			& $93.0$ \\
aug. cubes + one-class SVM 								& $84.2$			& $93.4$ \\
aug. cubes + k-means + one-class SVM 				& $86.4$			& $93.7$ \\
aug. cubes + k-means + 1-NN								& $78.8$			& $91.5$ \\
\hline
NNC (ours)															& $88.9$			& $94.1$ \\
\hline
\end{tabular}
\end{center}
\vspace*{-0.1cm}
\caption{Abnormal event detection results (in $\%$) in terms of frame-level and pixel-level AUC on the Avenue data set. Our framework and its preliminary (ablated) versions are compared with several state-of-the-art approaches~\cite{Giorno-ECCV-2016,Hasan-CVPR-2016,Ionescu-ICCV-2017,Liu-CVPR-2018,Lu-ICCV-2013,Luo-ICCV-2017,Smeureanu-ICIAP-2017}, which are listed in temporal order.\label{tab_Avenue}}
}
\vspace*{-0.3cm}
\end{table}

We first compare our approach with several state-of-the-art approaches~\cite{Giorno-ECCV-2016,Hasan-CVPR-2016,Ionescu-ICCV-2017,Liu-CVPR-2018,Lu-ICCV-2013,Luo-ICCV-2017,Smeureanu-ICIAP-2017} that reported results on the Avenue data set. The corresponding frame-level and pixel-level AUC metrics are presented in Table~\ref{tab_Avenue}. The table also includes ablation results for stripped-down versions of our approach, to show the performance gain brought by each component. A basic approach based on spatio-temporal cubes and one-class SVM yields a frame-level AUC of $81.3\%$. When we augment the cubes, the frame-level AUC grows by $2.9\%$ (up to $84.2\%$). Another $2.2\%$ (up to $86.4\%$) are gained when we employ k-means and train one-class SVM models on all $k$ clusters. By removing the smaller clusters, we obtain an improvement of $2.5\%$ and reach a frame-level AUC of $88.9\%$. We also tested an approach that removes the smaller k-means clusters in the first stage, but replaces the one-class SVM in the second stage with a one nearest neighbor (1-NN) model based the Euclidean distance to the nearest cluster centroid. The obtained frame-level AUC is $78.8\%$, which is nearly $10\%$ lower than the result of NNC. This ablation result shows the importance of using one-class SVM after k-means in order to learn a tight border around each cluster.

Using Narrowed Normality Clusters of augmented spatio-temporal cubes, we are able to surpass the results reported in previous works in terms of frame-level and pixel-level AUC. Compared to the most recent works~\cite{Ionescu-ICCV-2017,Liu-CVPR-2018,Luo-ICCV-2017,Smeureanu-ICIAP-2017}, our framework brings an improvement of more than $3.8\%$ in terms of frame-level AUC. 
Since our framework is able to process the video online on a single CPU, we consider that our results on the Avenue data set are noteworthy.

\begin{table}[t]
\small{
\begin{center}
\begin{tabular}{|l|c|c|}
\hline
Method 															& Frame AUC    	& Pixel AUC 	\\
\hline
\hline
Hinami et al.~\cite{Hinami-ICCV-2017}				& $89.8$			& - \\
\hline
NNC (ours)	 													& $91.1$			& $94.3$ \\
\hline
\end{tabular}
\end{center}
\vspace*{-0.1cm}
\caption{Abnormal event detection results (in $\%$) in terms of frame-level and pixel-level AUC on the Avenue17 data set. Our framework is compared with~\cite{Hinami-ICCV-2017}.\label{tab_Avenue17}}
}
\vspace*{-0.1cm}
\end{table}

\begin{figure}[!t]
\begin{center}
\includegraphics[width=0.98\columnwidth]{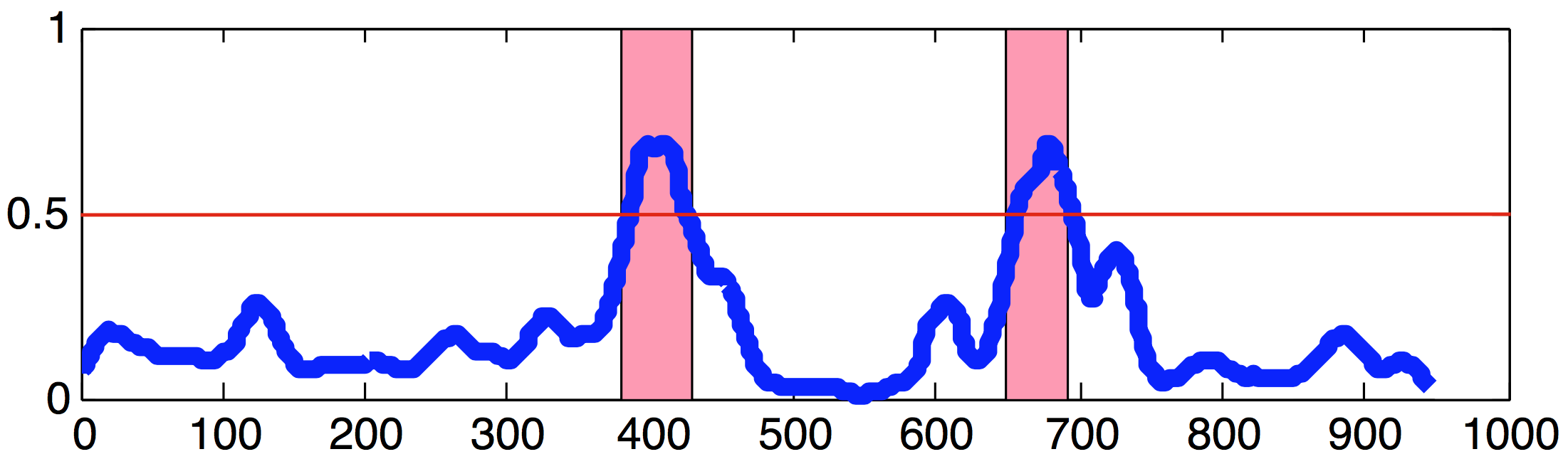}
\end{center}
\vspace*{-0.3cm}
\caption{Frame-level anomaly detection scores (between $0$ and $1$) provided by our approach for test video 4 in the Avenue data set. The video has $947$ frames. Ground-truth abnormal events are represented in pink and our scores are illustrated in blue. Best viewed in color.}
\label{fig_Avenue_vid4}
\end{figure}

\begin{figure}[!t]
\begin{center}
\includegraphics[width=0.9\columnwidth]{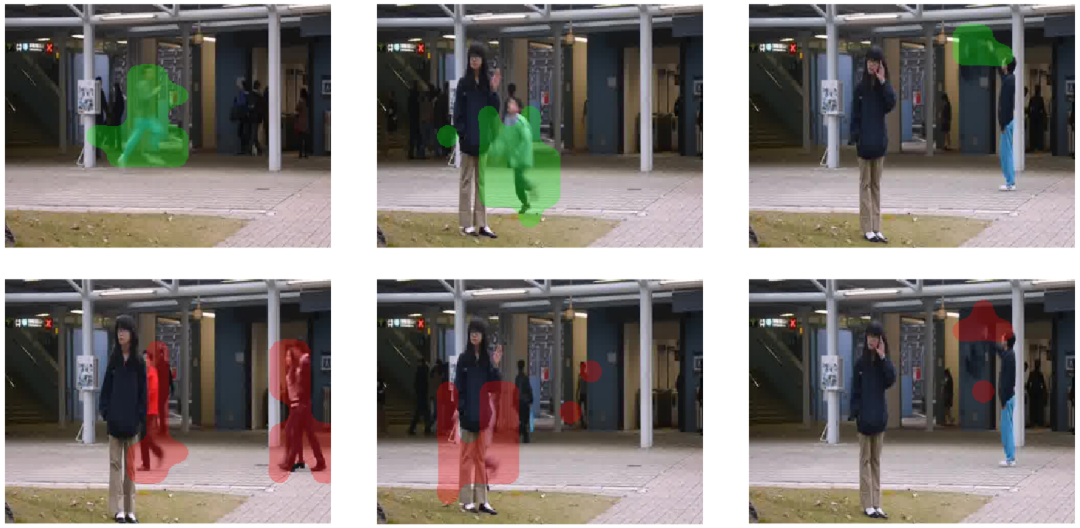}
\end{center}
\vspace*{-0.3cm}
\caption{True positive (top row) versus false positive (bottom row) detections of our framework. Examples are selected from the Avenue data set. Best viewed in color.}
\label{fig_Avenue_pos_neg}
\vspace*{-0.4cm}
\end{figure}

We also compare our approach with~\cite{Hinami-ICCV-2017} on the Avenue17 data set, a subset of the Avenue data set. Our frame-level AUC scores presented in Table~\ref{tab_Avenue17} are better than those reported by Hinami et al.~\cite{Hinami-ICCV-2017}. It is worth nothing that our approach yields better performance on the Avenue17 subset than on the full Avenue data set, indicating that the five removed test videos are actually more difficult than those left in Avenue17. As Hinami et al.~\cite{Hinami-ICCV-2017} observed, the removed videos contain abnormal objects that are not properly annotated, hence methods are prone to reach higher false positive rates on these five test videos.

Figure~\ref{fig_Avenue_vid4} depicts the frame-level anomaly scores produced by our approach against the ground-truth labels on test video 4 of the Avenue data set. We notice that our scores correlate well with the ground-truth labels. There are two abnormal events in this video and we can easily identify both of them by setting a threshold of $0.5$, without including any false positive detections. We also show some examples of true positive and false positive detections in Figure~\ref{fig_Avenue_pos_neg}. The true positive abnormal events are (from left to right) \emph{a person running}, \emph{a child running} and \emph{a person throwing an object}. The first (left-most) false positive detection represents \emph{two people walking synchronously}. The last two false positive examples indicate that our method detects \emph{a child running} even if the child is partially occluded, or \emph{a person throwing an object} before the object is in the air.

\vspace*{-0.1cm}
\subsection{Results on the Subway Data Set}
\vspace*{-0.1cm}

\begin{table}[t]
\small{
\begin{center}
\begin{tabular}{|l|c|c|c|}
\hline
Method 																& \multicolumn{3}{|c|}{Frame AUC} \\
\cline{2-4}
			 																& Entrance gate  	& Exit gate 	& Average\\
\hline
\hline
Cong et al.~\cite{Cong-CVPR-2011}						& $80.0$				& $83.0$ 		& $81.5$\\
Saligrama et al.~\cite{Saligrama-CVPR-2012}		& $89.1$	 			& - 				& -\\
Cheng et al.~\cite{Cheng-CVPR-2015}					& $92.7$ 				& - 				& -\\
Hasan et al.~\cite{Hasan-CVPR-2016}					& $94.3$				& $80.7$	 	& $87.5$\\
Del Giorno et al.~\cite{Giorno-ECCV-2016}			& $69.1$				& $82.4$ 		& $75.8$\\
Ionescu et al.~\cite{Ionescu-ICCV-2017}				& $71.3$				& $86.3$ 		& $78.8$\\
\hline
NNC (ours)															& $93.5$				& $95.1$ 		& $94.3$\\
\hline
\end{tabular}
\end{center}
\vspace*{-0.1cm}
\caption{Abnormal event detection results (in $\%$) in terms of frame-level AUC on the Subway data set. Our framework is compared with several state-of-the-art approaches~\cite{Cheng-CVPR-2015,Cong-CVPR-2011,Giorno-ECCV-2016,Hasan-CVPR-2016,Ionescu-ICCV-2017,Saligrama-CVPR-2012}, which are listed in temporal order.\label{tab_Subway}}
}
\vspace*{-0.1cm}
\end{table}

\begin{figure}[!t]
\begin{center}
\includegraphics[width=0.9\columnwidth]{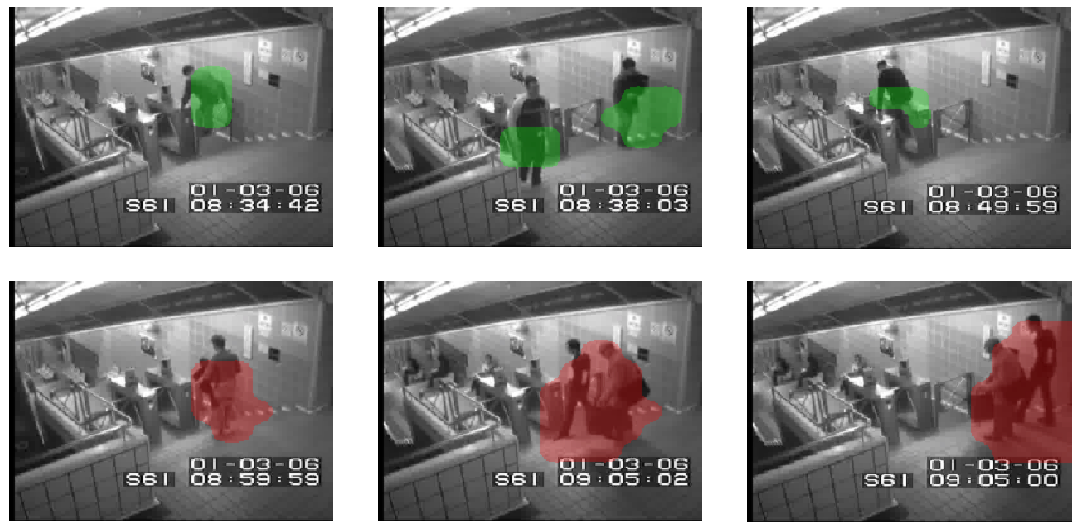}
\end{center}
\vspace*{-0.3cm}
\caption{True positive (top row) versus false positive (bottom row) detections of our framework. Examples are selected from the Subway Entrance gate. Best viewed in color.}
\label{fig_Subway_pos_neg}
\vspace*{-0.4cm}
\end{figure}

Although there are many works~\cite{Cheng-CVPR-2015,Cong-CVPR-2011,Giorno-ECCV-2016,Dutta-AAAI-2015,Hasan-CVPR-2016,Ionescu-ICCV-2017,Lu-ICCV-2013,Saligrama-CVPR-2012,Zhang-PR-2016} that report results on the Subway data set, some of these works~\cite{Dutta-AAAI-2015,Lu-ICCV-2013,Zhang-PR-2016} did not use the frame-level AUC as evaluation metric. Therefore, we only compare our approach with those methods~\cite{Cheng-CVPR-2015,Cong-CVPR-2011,Giorno-ECCV-2016,Hasan-CVPR-2016,Ionescu-ICCV-2017,Saligrama-CVPR-2012} that reported the frame-level AUC. The results of our comparative study are reported in Table~\ref{tab_Subway}. On the Entrance gate video, we obtain the second-best score, after Hasan et al.~\cite{Hasan-CVPR-2016}. They report a frame-level AUC of $94.3\%$, which is $0.8\%$ higher than our score ($93.5\%$). Things look differently on the Exit gate video, as we obtain the best score ($95.1\%$) among all methods, surpassing the approach of Hasan et al.~\cite{Hasan-CVPR-2016} by $14.4\%$ and the second-best method~\cite{Ionescu-ICCV-2017} by $8.8\%$. On average (last column in Table~\ref{tab_Subway}), we obtain the best results on the Subway data set.

In Figure~\ref{fig_Subway_pos_neg}, we present some interesting qualitative results obtained by our framework on the Entrance gate video. The true positive abnormal events are \emph{a person jumping over the fence}, \emph{two people walking in the wrong direction} and \emph{a person jumping over the gate}, while false positive detections are \emph{a person running} and \emph{two people walking synchronously}. 

\vspace*{-0.1cm}
\subsection{Results on the UMN Data Set}
\vspace*{-0.1cm}

\begin{table}[t]
\small{
\begin{center}
\begin{tabular}{|l|c|c|c|c|}
\hline
Method 																& \multicolumn{4}{|c|}{Frame AUC} \\
\cline{2-5}
			 																& \multicolumn{3}{|c|}{Scene} 							& All \\
\cline{2-4}
			 																& 1   					& 2 					& 3 					& scenes \\
\hline
\hline
Mehran et al.~\cite{Mehran-CVPR-2009}				& -					& - 					& -					& $96.0$ \\
Cong et al.~\cite{Cong-CVPR-2011}						& $99.5$			& $97.5$ 			& $96.4$			& $97.8$ \\
Saligrama et al.~\cite{Saligrama-CVPR-2012}		& -					& -	 				& -					& $98.5$ \\
Del Giorno et al.~\cite{Giorno-ECCV-2016}			& -					& -					& - 					& $91.0$ \\
Zhang et al.~\cite{Zhang-PR-2016}						& $99.2$			& $98.3$ 			& $98.7$			& $98.7$ \\
Sun et al.~\cite{Sun-PR-2017}								& $99.8$			& $99.3$ 			& $99.9$			& $99.7$ \\
Smeureanu et al.~\cite{Smeureanu-ICIAP-2017}	& $98.8$			& $93.6$ 			& $98.9$			& $97.1$ \\
Ionescu et al.~\cite{Ionescu-ICCV-2017}				& $99.3$			& $87.7$ 			& $98.2$			& $95.1$	 \\
Ravanbakhsh et al.~\cite{Ravanbakhsh-ICIP-2017}	& -				& -					& - 					& $99.0$ \\
\hline
NNC (ours)	 														& $99.9$			& $98.2$ 		& $99.8$ 				& $99.3$ \\
\hline
\end{tabular}
\end{center}
\vspace*{-0.1cm}
\caption{Abnormal event detection results (in $\%$) in terms of frame-level AUC on the UMN data set. Our framework is compared with several state-of-the-art methods~\cite{Cong-CVPR-2011,Giorno-ECCV-2016,Ionescu-ICCV-2017,Mehran-CVPR-2009,Ravanbakhsh-ICIP-2017,Saligrama-CVPR-2012,Smeureanu-ICIAP-2017,Sun-PR-2017,Zhang-PR-2016}, which are listed in temporal order.\label{tab_UMN}}
}
\vspace*{-0.1cm}
\end{table}

\begin{figure}[!t]
\begin{center}
\includegraphics[width=1.0\columnwidth]{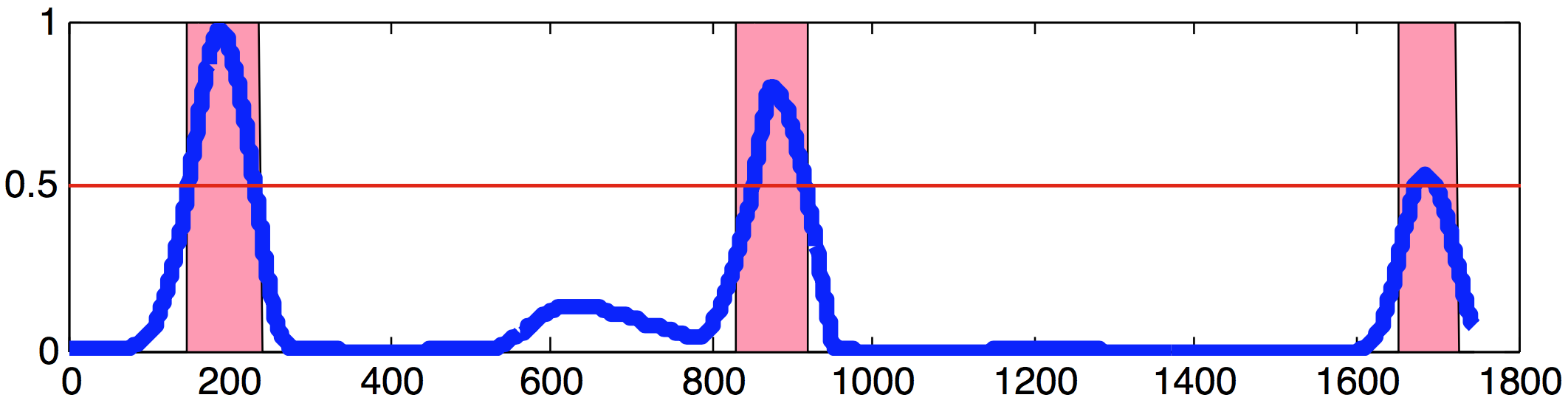}
\end{center}
\vspace*{-0.3cm}
\caption{Frame-level anomaly detection scores (between $0$ and $1$) provided by our framework for the third scene in the UMN data set. The video has $1744$ test frames. Ground-truth abnormal events are represented in pink and our scores are illustrated in blue. Best viewed in color.}
\label{fig_UMN_scene3}
\vspace*{-0.4cm}
\end{figure}

On the UMN data set, we compare our approach with several methods~\cite{Cong-CVPR-2011,Giorno-ECCV-2016,Ionescu-ICCV-2017,Mehran-CVPR-2009,Ravanbakhsh-ICIP-2017,Saligrama-CVPR-2012,Smeureanu-ICIAP-2017,Sun-PR-2017,Zhang-PR-2016}. In Table~\ref{tab_UMN}, we report the frame-level AUC score for each individual scene, as well as the average score for all the three scenes. It is worth noting that UMN seems to be the easiest abnormal event detection data set, since most works report frame-level AUC scores above $95.0\%$. We reach the highest performance ($99.9\%$) among all methods on the first scene. On the last scene, we obtain the second best score ($99.8\%$). Remarkably, our approach is able to correctly identify the three abnormal events in the third scene without any false positives, by applying a threshold of $0.5$, as illustrated in Figure~\ref{fig_UMN_scene3}. Our lowest performance ($98.2\%$) is on the second scene. Over all scenes, we reach the second highest frame-level AUC ($99.3\%$), which is $0.4\%$ lower than the best score ($99.7\%$) obtained by Sun et al.~\cite{Sun-PR-2017}.

\begin{figure}[!t]
\begin{center}
\includegraphics[width=0.9\columnwidth]{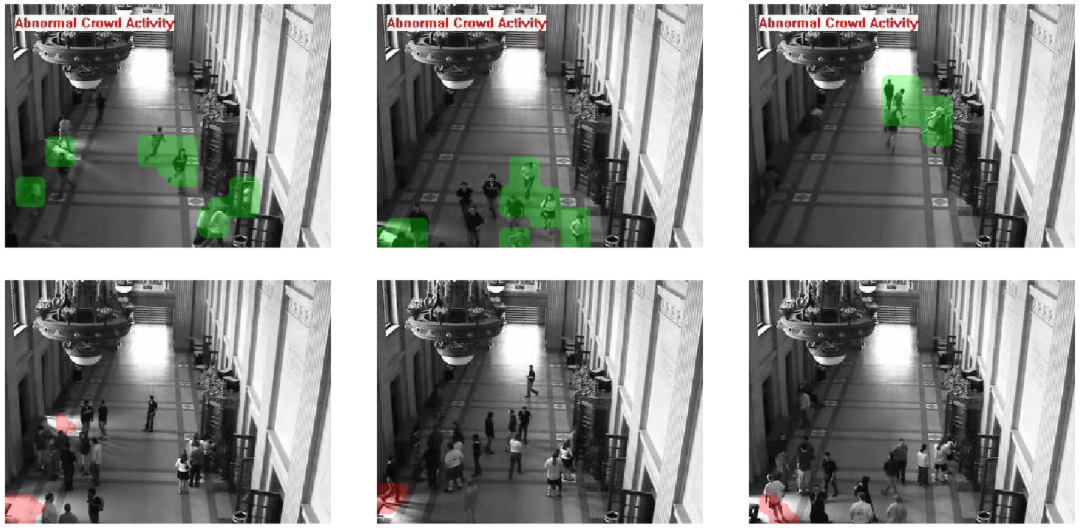}
\end{center}
\vspace*{-0.3cm}
\caption{True positive (top row) versus false positive (bottom row) detections of our framework. Examples are selected from the UMN data set. Best viewed in color.}
\label{fig_UMN_pos_neg}
\vspace*{-0.4cm}
\end{figure}

In Figure~\ref{fig_UMN_pos_neg}, we present some interesting qualitative results obtained by our framework on the second scene, as it was almost impossible to find false positive detections in the other scenes. The true positive examples represent \emph{people running around in all directions}, while the false detections are triggered by \emph{people opening the doors} to enter or exit the room. In the first (left-most) false positive example, it seems that our method detects the significant amount of light that enters the room as the doors open. Perhaps the first impression is that our approach is not robust to illumination variations. However, we noticed that our training video does not contain examples of people walking through the doors. Therefore, it is impossible to learn a complete model of normality that includes this kind of event \emph{(people walking through the doors)}.

\vspace*{-0.2cm}
\section{Conclusion and Future Work}
\label{sec_Conclusion}
\vspace*{-0.1cm}

In this work, we proposed \emph{Narrowed Normality Clusters}, a novel framework for abnormal event detection in video that is based on a two-stage outlier elimination algorithm. The algorithm works by removing outlier clusters obtained with k-means and by learning a tight border around each remaining cluster using one-class SVM. Our secondary contribution was to augment the spatio-temporal cubes with location, motion direction, and deep appearance features.
We conducted abnormal event detection experiments on three data sets to compare our approach with a series of state-of-the-art approaches~\cite{Cheng-CVPR-2015,Cong-CVPR-2011,Giorno-ECCV-2016,Hasan-CVPR-2016,Hinami-ICCV-2017,Ionescu-ICCV-2017,Liu-CVPR-2018,Lu-ICCV-2013,Luo-ICCV-2017,Mehran-CVPR-2009,Ravanbakhsh-ICIP-2017,Saligrama-CVPR-2012,Smeureanu-ICIAP-2017,Sun-PR-2017,Zhang-PR-2016}. The empirical results indicate that our approach yields better performance than all other methods on the Avenue and the Subway data sets. Furthermore, our approach is second best on the UMN data set. At the same time, we can process the test video in real-time at $24$ frames per second on a single CPU. In future work, we aim to develop an approach to train deep features on a closely related task, such as action recognition, and transfer the learned features to our task.

\vspace*{0.3cm}
\noindent
{\bf Acknowledgments.}
The work of Radu Tudor Ionescu was supported by a grant of Ministery of Research and Innovation, CNCS - UEFISCDI, project number PN-III-P1-1.1-PD-2016-0787, within PNCDI III.

{\small
\bibliographystyle{ieee}
\bibliography{references}
}

\end{document}